\documentclass[sigconf]{acmart}

\usepackage[utf8]{inputenc} 
\usepackage[T1]{fontenc}    
\usepackage{hyperref}       
\usepackage{url}            
\usepackage{booktabs}       
\usepackage{amsfonts}       
\usepackage{nicefrac}       
\usepackage{microtype}      
\usepackage{xcolor}         
\usepackage{color}
\usepackage{cases}

\usepackage{makecell}
\usepackage{multirow}
\usepackage{microtype}
\usepackage{booktabs}
\usepackage{enumitem}
\usepackage[utf8]{inputenc}
\usepackage{listings}
\usepackage{subcaption}
\usepackage{dsfont}
\usepackage{graphicx}
\usepackage[linesnumbered,ruled,lined]{algorithm2e}

\usepackage{ulem}
\usepackage{afterpage}
\usepackage{xspace}
\usepackage{adjustbox}

\newcommand{\ie}{\textit{i.e.}\xspace} 
\newcommand{\eg}{\textit{e.g.}\xspace} 
\newcommand{\etc}{\textit{etc.}\xspace} 

\newcommand{\xhdr}[1]{\vspace{1.7mm}\noindent{{\bf #1}}}

\newtheorem{thm:eg}{Example}
\def \E {\mathrm{E}}

\def \A {\mathcal{A}}

\def \E {\mathcal{E}}

\def \G {\mathcal{G}}

\def \S {\mathcal{S}}
\def \T {\mathcal{T}}
\def \V {\mathcal{V}}

\def \X {\mathbf{X}}

\newcommand{\Ours}{\textsc{GPEFT}\xspace}

\usepackage[disable,textsize=tiny]{todonotes}

\AtBeginDocument{%
  \providecommand\BibTeX{{%
    \normalfont B\kern-0.5em{\scshape i\kern-0.25em b}\kern-0.8em\TeX}}}

\begin{document}
\title{Parameter-Efficient Tuning Large Language Models for Graph Representation Learning}

\author{Qi Zhu}
\affiliation{%
  \institution{Amazon Web Services}
  \country{}
}
\email{qzhuamzn@amazon.com}

\author{Da Zheng}
\affiliation{%
  \institution{Amazon Web Services}
  \country{}}
\email{dzzhen@amazon.com}

\author{Xiang Song}
\affiliation{%
  \institution{Amazon Web Services}
  \country{}
}
\email{dzzhen@amazon.com}

\author{Shichang Zhang}
\authornote{Work done while being an intern at Amazon.}
\affiliation{%
 \institution{University of California, Los Angeles}
 \country{}}
\email{shichang@cs.ucla.edu}

\author{Bowen Jin}
\authornotemark[1]
\affiliation{%
  \institution{University of Illinois Urbana-Champaign}
  \country{}}
\email{bowenj4@illinois.edu}

\author{Yizhou Sun}
\affiliation{%
  \institution{University of California, Los Angeles}
  \country{}}
\email{yzsun@cs.ucla.edu}

\author{George Karypis}
\affiliation{%
  \institution{Amazon Web Services}
  \country{}}
\email{gkarypis@amazon.com}

\begin{abstract}
Text-rich graphs, which exhibit rich textual information on nodes and edges, are prevalent across a wide range of real-world business applications.
Large Language Models (LLMs) have demonstrated remarkable abilities in understanding text, which also introduced the potential for more expressive modeling in text-rich graphs.
Despite these capabilities, efficiently applying LLMs to representation learning on graphs presents significant challenges.
Recently, parameter-efficient fine-tuning methods for LLMs have enabled efficient new task generalization with minimal time and memory consumption.
Inspired by this, we introduce Graph-aware Parameter-Efficient Fine-Tuning - \Ours, a novel approach for efficient graph representation learning with LLMs on text-rich graphs. 
Specifically, we utilize a graph neural network (GNN) to encode structural information from neighboring nodes into a graph prompt. This prompt is then inserted at the beginning of the text sequence.
To improve the quality of graph prompts, we pre-trained the GNN to assist the frozen LLM in predicting the next token in the node text.
Compared with existing joint GNN and LMs, our method directly generate the node embeddings from large language models with an affordable fine-tuning cost.
We validate our approach through comprehensive experiments conducted on 8 different text-rich graphs, observing an average improvement of 2\% in hit@1 and Mean Reciprocal Rank (MRR) in link prediction evaluations. Our results demonstrate the efficacy and efficiency of our model, showing that it can be smoothly integrated with various large language models, including OPT, LLaMA and Falcon.
\end{abstract}




\maketitle

\section{Introduction}\label{sec:introduction}
Information networks form the backbone of modern data systems: millions of daily social posts are shared among users on platforms like Facebook and Twitter; countless papers are published and cited within academic networks. Many of these networks are rich in textual information on various types of objects, known as text-rich or text-attributed networks. For example, in an e-commerce graph, text-rich information could be used to predict links by analyzing product descriptions, user reviews, and metadata to recommend items that are frequently bought together. 
Such real-world applications typify the problem of link prediction, where representation learning (\ie, embedding) emerges as the most prevalent solution.

Representation learning on graph-structured data aims to learn a dense vector for each node through self-supervised tasks such as link prediction~\cite{perozzi2014deepwalk,grover2016node2vec,zhang2019heterogeneous} or masked attribute recovery~\cite{hou2022graphmae}. 
These dense vectors can be widely utilized in various downstream applications including search, ranking and retrieval. 
To effectively utilize both node attributes and structural information in representation learning, Graph Neural Networks (GNNs)~\cite{gcn,graphsage} devise a novel type of neural networks with message-passing mechanism.
In text-rich graphs, raw text is usually transformed into feature vectors by a text encoder prior to the application of Graph Neural Networks (GNNs). 
Early text encoders, such as bag-of-words and Word2Vec\cite{word2vec} , are being gradually phased out due to significant improvements in text representation from transformer-based language models (LMs). 


This has spurred interest in jointly modeling textual and structural data with GNNs and LMs. Notable architectures include (1) a cascading architecture that integrates GNNs with LM features (cascading GNN-LMs); (2) nested GNN-LMs that blend message passing into the language modeling process. 
Examples of the latter include GraphFormer~\cite{yang2021graphformers} and Heterformer~\cite{jin2022heterformer}, which introduce a nested GNN-transformer architecture for homogeneous and heterogeneous graph, respectively.
Historically, GNN-LMs have focused on combining GNNs with medium-sized LMs such as BERT~\cite{devlin2018bert} and RoBERTa~\cite{liu2019roberta}. 
However, large language models (LLMs)~\cite{brown2020language,touvron2023llama,zhang2022opt}  with billions of parameters, known for their exceptional multi-task generalization and instruction-following capabilities, present an untapped potential for further enhancing the performance of these models.
Applying large language models (LLMs) to text-rich graphs presents an intriguing concept; however, fine-tuning such GNN-LLM models, as with existing architecures, is impractical due to the prohibitively high computational and memory demands.
As an alternative, parameter-efficient fine-tuning (PEFT) techniques for LLMs, such as LORA~\cite{hu2021lora} and Prefix Tuning~\cite{li2021prefix}, advocate for updating a minimal fraction of parameters (\eg, less than 1\%) to achieve comparable performance.
Inspired by this, our method integrates graph encoding into the PEFT framework, which enables efficient graph representation learning with LLMs.

Contrary to PEFT, another line of research~\cite{ye2023natural,fatemi2023talk} explores in-context learning of LLMs without weight updates. These works transform the neighboring context of a target node into textual descriptions, incorporating features, and employ LLMs to make predictions such as node degrees, labels, and the existence of links.
However, applying these methods to representation learning at an industrial scale presents notable challenges. These approaches are primarily tailored for tasks involving a limited number of candidates and do not account for the similarity between any pair of nodes. In a graph with $N$ nodes, they require a substantial inference cost of $\mathcal{O}(N^2)$ to iterate every positive edge, in contrast to the more efficient strategy of generating node embeddings at a cost of $\mathcal{O}(N)$ and subsequently calculating their similarities within the embedding space (see also the inference cost of in-context learning in Table~\ref{tab:method_comparison}).
This disparity shows the limitation of in-context learning for text-rich graphs and emphasizes the need for more scalable methods to handle large-scale graph data efficiently.



\begin{figure}[t]
\begin{center}
\includegraphics[width=1\columnwidth]{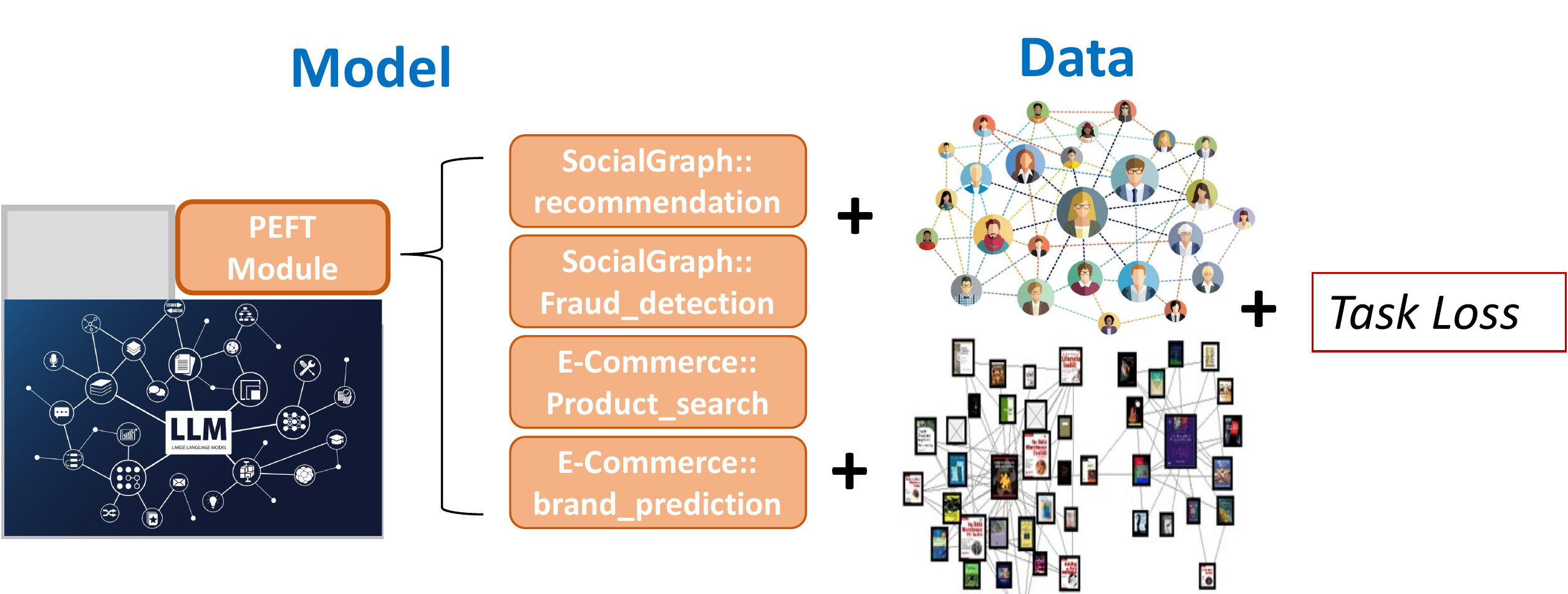}
\end{center}
\caption{Convenient use case of parameter-efficient fine-tuning for text-rich graphs, requiring only 2\% additional LLM parameters per graph application.}
\label{fig:teaser}
\end{figure}

In this work, we introduce Graph-aware Parameter-Efficient Fine-Tuning (\ie, \Ours) for \textbf{light-weighted} and \textbf{efficient} graph representation learning with LLMs. 
Our methodology leverages graph neural networks (GNNs) to encode neighboring nodes into a graph prompt, which is subsequently integrated into the original sequence as a ``soft prompt''. We employ a pre-trained LLM and update only the GNN prompt encoder and the PEFT parameters, ensuring minimal computational cost.
As shown in Figure~\ref{fig:teaser}, \Ours does not change the original parameters of LLMs and only $\sim 2\%$ of additional parameters are optimized for each task on a given text-rich graphs. For instance, for recommendations on social networks, we can simply load the SocialGraph::recommendation component into the backbone LLM and train \Ours with task-specific supervision.
To improve the quality of the graph prompt, we introduce a pre-training phase for the GNN prompt encoder. This phase utilizes a next-token prediction objective on node texts to align the distribution of graph prompt and word embeddings of LLMs.
In the fine-tuning (\ie, representation learning) phase, we choose to use contrastive loss~\cite{hadsell2006dimensionality} with one negative sample to avoid the memory overhead for traditional in-batch loss used in LM fine-tuning~\cite{gao2021simcse}. 

Compared with existing GNN-LMs, we reduce the training cost of GNN-LMs by parameter-efficient tuning and use structural information as input instead of a nested architecture; Compared with In-Graph Context Learning LLMs, we avoid the quadratic inference costs for link prediction. In Table~\ref{tab:method_comparison}, we calculate the approximated computation cost for these two methods and \Ours.
In our experiments, we conduct large-scale link prediction evaluation through representation learning across 8 different graphs in two domains. Our observations reveal an average improvement of 2\% over large language models (LLMs) and 3\% over the existing GNN-LMs.
Our contributions are summarized as follows,
\begin{itemize}
\item We propose the first graph representation learning framework that utilizes Large Language Models with billions of parameters.
\item We develop a novel Parameter-Efficient Fine-Tuning (PEFT) algorithm - \Ours for text-rich graphs. This method effectively integrates the key structural information into LLMs as a graph prompt.
\item Through comprehensive experiments conducted on eight text-rich graphs, we consistently observe improvements with our approach over previous GNN-LMs, while also achieving computational efficiency.
\end{itemize}

\begin{table}[t]
\centering

\caption{Comparison of existing GNN-LMs for representation learning on a graph with one million edges. We use GPU hours to estimate the computation cost.}
\label{tab:method_comparison}
\begin{tabular}{ c  c  c  c }
 \toprule
  & GNN-LM~\cite{jin2023patton} &  LLM Service~\cite{ye2023natural} & \Ours \\
 \midrule
  category & fine-tune & in-context learning & PEFT\\
 \#params per task & 110M & - & 14.4M\\
training cost & $\sim$\$250 & - & $\sim$\$50\\
prediction cost & $\sim$\$5 &  $>$\$1000 & $\sim$\$10 \\
\bottomrule
\end{tabular}
\end{table}

\section{Related Work}\label{sec:related}
\xhdr{Modeling Text-Rich Graphs using LMs.}
Regarding the success of language models and GNNs in their respective areas, modeling text-rich graphs has been a hot topic\cite{jin2023large}. \textbf{(1) Graph-empowered GNN-LMs}: GraphFormers~\cite{yang2021graphformers} is a GNN-nested Transformer architecture that insert GNN modules between transformer layers. 
Using language models to model target and neighbor node texts requires huge memory and time costs. To address this, some work \cite{liu2019fine} proposed freezing the language model to reduce the computation needed for cascading. 
Some work \cite{li2021adsgnn, jin2022heterformer} proposed neighbor sampling but that reduces the graph information captured.
Therefore, recently some work propose to joint train LMs and GNNs through knowledge distillation \cite{mavromatis2023train} or Expectation Maximization algorithms \cite{zhao2022learning}.
\textbf{(2) Self-supervised GNN-LMs}: some methods \cite{chien2021node, mavromatis2023train} directly supervise language model fine-tuning through graph-related tasks, to help language models better understand the textual information in text-attributed graphs. The language model is then combined with GNNs by freezing the language model. This approach demonstrates the inherent connections between graph structure and text in TAGs. However, current research in this direction has limitations in that the LM and graph are separate, and the language model cannot directly perceive graph information. It also does not utilize the inherent connections between language and graphs to help GNNs better learn structural features. \textbf{(3) LLMs for Graph}: With the breakthrough progress made by LLMs on textual tasks \cite{touvron2023llama, brown2020language}, recently many works have emerged exploring how to directly utilize LLMs to understand text-attributed graphs \cite{chen2023exploring}.
Some works also explored using large models to enhance the textual features of text-attributed graphs \cite{he2023harnessing, duan2023simteg}. In terms of the representation learning, methods~\cite{fatemi2023talk,tian2023graph,ye2023natural} that turns graph structure into in-context learning are most viable option for representation learning.

\xhdr{Parameter-Efficient Fine-Tuning of LLMs.}
As the size of language model continue to increase, full fine-tuning has became more and more impractical. Parameter-efficient fine-tuning (PEFT)~\cite{peft} freeze most of the language model parameters, which has been successfully applied to popular language models such as BERT~\cite{devlin2018bert}, GPT~\cite{radford2018improving,radford2019language,brown2020language} and t5~\cite{2020t5}. This section reviews the key developments in the area. \textbf{(1) Adapter Layers~\cite{houlsby2019parameter}}: Adapter layers are small, trainable modules inserted between transformer layers of a pre-trained langauge model.  \textbf{(2) Prompt Tuning~\cite{lester2021power}}: Prompt tuning leverages the pre-trained knowledge of LLMs by appending task-specific prompts to the input text, effectively guiding the model's predictions without updating its parameters. \textbf{(3) Low-Rank Adaptation(LORA)~\cite{hu2021lora}}: LORA modifies the attention and feedforward layers in the transformer through two low-rank updating matrices. These approaches can also be combined and found to be useful in applications like multi-modal instruction tuning models~\cite{liu2023visual,gao2023llama}.

\section{Notations and Preliminaries} \label{sec:preliminary}
This section introduces the background knowledge, notations and technical terms that will be used throughout the paper.

\subsection{Text-Rich Graph Representation Learning}
Let $\G = (\V, \X, \E)$ be defined as a text-rich graph with multiple edge types $\T$, 
where $\V$ represents the set of nodes, $\X$ denotes the text features associated with nodes, and $\E$ symbolizes the edges between nodes. 
Each node $i$ is accompanied by a text sequence (\eg, item description, paper title and abstract) $\S_i = \{s_{i,0},...,s_{i,k}\}$. For each edge type $t \in \T$, an edge $e_{ij} \in \E_t$ indicates the presence of an edge between node $i$ and $j$ of type $t$. Given the target edge type $t$, we define the problem of representation learning as mapping nodes into the embedding space, considering all observed edges except those designated for testing, denoted by $\E \setminus \E_t^\text{test}$. The learning objective aims to maximize the likelihood of observing the training edges $\E_t^\text{train}$. The evaluation of the learned node representations is conducted through link prediction on the testing edges $\E_t^\text{test}$.

\subsection{Casual Language Modeling}
Given a text sequence on each node, the causal language model (CLM)~\cite{radford2018improving} aims to learn the probability distribution of these text sequences in a way that respects the sequential order of words or tokens. 
CLM is the most popular architecture for recent large language models such as GPT-3~\cite{brown2020language} and LLaMA~\cite{touvron2023llama}.
Specifically, the objective function of the CLM, $\mathcal{L}_\text{LLM}$ is defined as the negative log-likelihood of predicting each subsequent token in the sequence given its preceding tokens. This training objective is also referred as ``next token prediction'' frequently in the literature.
\begin{equation}
\mathcal{L}_\text{LLM} = - \sum_{i} \sum_{j=0}^{k} \log P(s_{i,j+1} | s_{i,0}, \ldots, s_{i,j}; \Theta_\text{LLM})
\end{equation}
In this equation,  $P(s_{i,j+1} | s_{i,0}, \ldots, s_{i,j}; \Theta_\text{LLM}$ represents the probability of generating the next token $s_{i,j+1}$ given the sequence of preceding tokens $(s_{i,0}, \ldots, s_{i,j})$, parameterized by $\Theta_\text{LLM}$. The architecture of a Large Language Model (LLM) is predominantly based on transformers~\cite{vaswani2017attention}. This work explores the potential of using casual lanugage model to learn the node representations in text-rich graphs.

\xhdr{Prompt Tuning.} 
A prompt for LLMs is a text snippet concatenated with the original text sequence designed to bridge the gap between pre-training tasks (e.g., CLM) and downstream tasks, such as question answering or summarization.
In prompt tuning, soft prompt denotes a method where learnable embeddings are utilized to steer the model's response ~\cite{lester2021power, zhong2021factual, liu2022p} instead of supplying a concrete textual prompt. These soft prompts are typically concatenated with the input text data. Unlike fixed textual prompts, soft prompts are adaptable and are optimized to elicit the most effective response from the model. We model the structural information in $\G$ as a soft prompt in this work.

\subsection{Parameter-Efficient LLM Fine-tuning}
\begin{enumerate}
    \item Low-Rank Adaptation (LORA): Consider $W$ as the original weight matrix in attention and feed-forward layer of the transformer. LoRA introduces two low-rank matrices $A$ and $B$, where the product $AB^T$
  approximates the desired change in $W$. The modified weight matrix $\hat{W}$
  is then:
    \begin{equation}
    \hat{W} = W + AB^T
    \end{equation} 
    In this equation, the matrices A and B are trainable, while W remains fixed. The low-rank nature of $AB^T$ ensures that the number of additional parameters is relatively small. 
    \item Prefix-Tuning: Prefix Tuning involves adding a small number of trainable parameters (prefixes) to the input of each transformer layer. Let's denote the prefix as P and the original input as $\S$. The modified input to the transformer layer can be represented as: $[P;\S]$.
    
\end{enumerate}

\section{Method}\label{sec:method}
This section presents our proposed framework - \Ours, which employs graph-aware PEFT techniques for representation learning in Large Language Models (LLMs). The core idea of \Ours includes two key strategies: (1) integrating graph information through parameter-efficient tuning methods in LLMs, and (2) executing large-scale graph representation learning specifically for the task of link prediction.

\begin{figure*}[t]
\begin{center}
\includegraphics[width=1\textwidth]{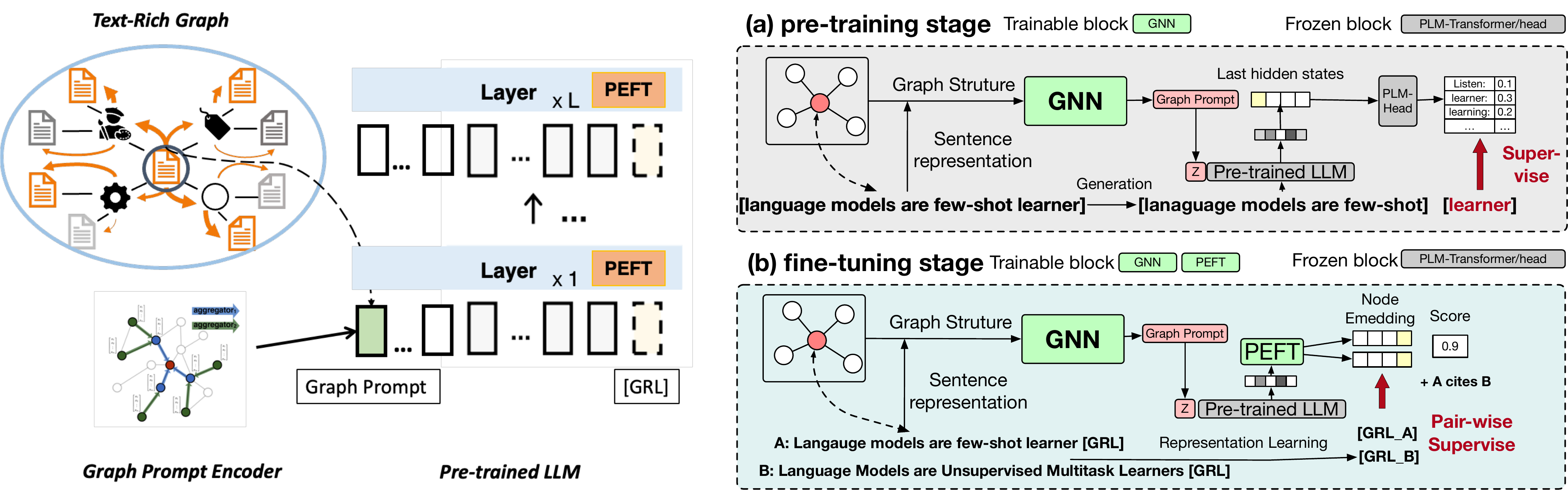}
\end{center}
\caption{Overview of the \Ours Framework: Architecture (Left) and Pre-training and Fine-tuning Processes (Right).}
\label{figure:framework}
\end{figure*}

\xhdr{Motivation:} Language models accurately encode the textual information in a text-rich graph, many structure-related predictions, however, are difficult to infer from the text alone. As illustrated in Figure~\ref{figure:framework}, some popular items in an e-commerce network or seminal papers in an academic graph are densely connected, even though their textual similarity to neighboring nodes varies significantly. Although several recent works~\cite{he2023harnessing,ye2023natural,fatemi2023talk} have integrated graph information into Large Language Models (LLMs), most focus on node or graph classification problems. However, the nature of link prediction differs significantly from classification, as it requires calculating the proximity between pairs of nodes. Motivated by this distinction, our work aims to learn node embeddings through graph-aware LLM fine-tuning.

\xhdr{Framework:} The framework of \Ours is shown in Figure~\ref{figure:framework}. Given a text-rich graph $\G$, and target edge type for training, there are following three steps for representation learning: step 1: a GNN model called graph prompt encoder \textbf{prepend} the graph structural information before text sequence in the language model, we detail the descriptions in Section~\ref{subsec:gnn_encoder}; step 2: 
To align the feature space of node representations output by GNN with the text embeddings in a LLM, we consider to use the casual language modeling to \textbf{pre-train} the GNN prompt encoder; step 3: an computationally efficient contrastive learning loss is employed to optimized the GNN-LLM model end-to-end with \textbf{parameter-efficient fine-tuning} techniques. We describe the pre-training and fine-tuning details in subsections of Section~\ref{subsec:train_gnnllm}.

\subsection{Graph prompt tuning}\label{subsec:gnn_encoder}
In the existing graph LLMs~\cite{fatemi2023talk, ye2023natural}, the graph encoding function $\Theta_g$ typically transforms the neighbors of node $i$ into a textual sequence, such as ``{node\_1: text, node\_2: text}, node\_1 is connected with node\_2...''. 
However, this approach assumes that all neighbors are equally important to the target node, which can also result in very long input sequences for the LLM in a densely connected graph.

\xhdr{Structural representations.}
On the contrary, we transform the node texts $x_i$ and structural information $\A$ into dense vectors using GNNs. To achieve this, we fist use a small pre-trained langauge model - BERT~\cite{devlin2018bert} to turn the text sequence $\S_i$ into document embedding $x_i$. Then, we use a common message passing GNN like GCN~\cite{gcn} or GraphSAGE~\cite{graphsage} to obtain the GNN embedding $z_i$ for node $i$ in the given graph,
\begin{equation}\label{eq:gnn_encoder}
    z_i = \text{GNN}(x_i, \mathcal{X}, \mathcal{A} ; \Theta_g)
\end{equation}
where $x_i$ is the center node feature, $\mathcal{X}$ and $\mathcal{A}$ are neighbor node features and adjacency matrix, respectively. $\Theta_g$ represents the GNN parameters, which also
contains a mapping function from the GNN embedding space to the word embedding space of the LLM, $\mathcal{M}:\mathbb{R}^{d_{\text{GNN}}} \rightarrow \mathbb{R}^{d_{\text{model}}}$, and $d_{\text{GNN}} < d_{\text{model}}$.
We also refer $z_i$ as the \textbf{(node-level) graph prompt} in the remaining of the paper.

\xhdr{Prompting LLM with the graph prompt:} 
With the newly created graph-aware representation of node $z_i$, we now discuss two different soft prompting strategies: (1) prepend or (2) append graph prompt to the target node sequence $\{s_{i,0},...,s_{i,k}\}$. We denote the hidden states in the transformer for $j-th$ word in text sequence $\S_i$  as $h_{i,j}$. Following the prior research of representation learning using casual language model~\cite{muennighoff2022sgpt}, we introduce a specialized graph representation learning \texttt{[GRL]} token at the end of the sequence to represent the final node representations. We opt to \textbf{prepend} the graph prompt because it offers greater expressive power, as detailed in the following theorem.

\begin{theorem}[]\label{thm_1}
Consider the $l$-th layer of a transformer, where $H^l_k = [h^l_{i,0}, \ldots, h^l_{i,k}]$ represents the embedding matrix for a text sequence of length $k$. At each layer, appending a graph prompt to the sequence results in a convex combination of the embeddings $[H^l_k; z_i]$. In contrast, prepending the graph prompt $z_i$ to the sequence allows for the incorporation of $z_i$ in the computation of $H^l_k$.
\end{theorem}
We observe that in the attention calculation, prepending the graph prompt can potentially enhance the model's expressiveness through direct interaction with the text sequence's embeddings. The detailed proof is provided in Appendix A.



\subsection{Graph representation learning with LLMs}~\label{subsec:train_gnnllm}
Following the previous section, we compute the node representation $v_i$ as the last hidden representation of the special token \texttt{[GRL]} as follows:
\begin{equation}\label{eq:inference}
    v_i = \text{LLM}\left(\{ z_i, s_{i,0}, s_{i,k}, \texttt{[GRL]}\} ; \Theta_g, \Theta_\text{LLM}\right)
\end{equation}
where $\Theta_g$ and $\Theta_\text{LLM}$ are the trainable parameters of GNN and LLM. 
Now we discuss the motivation of pre-trainng $\text{GNN}$ through casual language modeling.

\xhdr{Pre-training GNN Prompt Encoder.} Note that node features $x_i$ are independent to the large language model and the GNN $\Theta_g$ is randomly initialized. The feature distribution of the GNN prompt encoder $P(z_i)$ does not align with the word embeddings $P(h^0_i)$. We use the next token prediction objective to pre-train the GNN \textcolor{blue}{$\Theta_g$} while freezing the LLM $\Theta_\text{LLM}$,
\begin{equation}\label{eq:pre-train}
\mathcal{L}_\text{pre-training} = - \sum_{i} \sum_{j=0}^{k} \log P(s_{i,j+1} | z_i, s_{i,0}, \ldots, s_{i,j}; \textcolor{blue}{\Theta_g}, \Theta_\text{LLM})
\end{equation}
where $\mathcal{L}$ is the loss function, $P(s_{i,j+1} | z_i, s_{i,0}, \ldots, s_{i,j}; \Theta)$ 
is the conditional probability of the token $s_{i,j+1}$ given the GNN prompt $z_i$ 
and the preceding tokens up to $s_{i,j}$, and \textcolor{blue}{$\Theta$} represents the trainable parameters of this step.

\xhdr{Fine-tuning.}
After pre-training, we use the cosine similarity between the \texttt{GRL} embeddings of node pairs to represent the likelihood of edges in the graph.
In particular, we obtain the representations of each node $v$, compute the cosine distance $d_{ij} = 1 - \cos (v_i, v_j)$ between node $i$ and $j$ and employ a contrastive loss~\cite{hadsell2006dimensionality} between a positive edge $e_{ij}$ and random negative edge $e_{ij'}$:
\begin{eqnarray}\label{eq:fine-tune}
    \mathcal{L}_\text{fine-tuning} = \sum_{e_{ij} \in \mathcal{E}} \left(d_{ij}^2 + \max(\tau - d_{ij'}, 0)^2 \right) \\
    v_i = \text{LLM}\left(\{ z_i, s_{i,0}, s_{i,k}, \texttt{[GRL]}\} ; \textcolor{blue}{\Theta_g}, \textcolor{blue}{\Theta_\text{LLM}}\right)
\end{eqnarray}
it encourages connected nodes to have similar representations while penalizes disconnected nodes when their cosine similarity is larger than the margin $\tau$ (\ie, $\tau=0.5$ in our experiment).
In a LLM with billion parameters, usually the batch size cannot be large, hence we observe using such loss yields better performance and computational efficiency than traditional in-batch contrastive learning loss (\eg SimCSE~\cite{gao2021simcse}) with a small batch size. Note that in this step, we update both the LLM and GNN parameters.

\subsection{Model Optimization.} 
We outlined the training process of our framework in the previous section. However, it is well-known that directly optimizing a billion-scale LLM is impractical. Furthermore, maintaining a different set of full LLM parameters for various types of text-rich graphs can lead to space inefficiency.
A more efficient approach would be the ability to store a small amount of parameters for a specific application related to a text-rich graph. To this end, we propose partitioning the parameters of the LLM into two segments: $\Theta_\text{LLM} = [\textcolor{blue}{\Theta_\text{peft}}; \Theta_\text{pre}]$, where $\Theta_\text{pre}$ indicates the pre-trained weights of a specific LLM. This division allows \textbf{light-weighted} parameters for each specific application.

Now we describes how we integrates our graph prompt encoder \textcolor{blue}{$\Theta_g$} into traditional parameter efficient tuning \textcolor{blue}{$\Theta_\text{peft}$} while maintaining the representation power discussed before. 
We consider use two most popular peft variants:
\begin{enumerate}
    \item Low-Rank Adaptation (LORA): In LORA, the trainable parameters are coupled in every transformer layer. Therefore, the naive prepending of the GNN prompt can effectively participate in the gradient updates of LORA matrices.
    \item Prefix-Tuning: In prefix-tuning, the modified input to the transformer layer is typically represented as: $[P; \S]$. However, placing graph tokens $Z$ at the beginning of $\S$ does not contribute to the updating of the prefix embedding $P$ similar to the Theorem~\ref{thm_1}. To address this, we propose a slight modification to the prefix embedding, denoted as $P' = P + Z$. This adjustment ensures that the graph information is effectively integrated with the prefix tuning process.
    
\end{enumerate}

\begin{algorithm}
\textbf{Input:} text-rich graph $\G$, a set of training edges $\E_t^\text{train}$\\
pre-trained LLM: $\Theta_{LLM}$, GNN encoder: $\Theta_g$, \\
Graph Neighborhood Sampler \textbf{\texttt{SAMPLE}}.\\
\textbf{Output:} node embeddings $V$ for all nodes \\ 
// Pre-training;\\
\For{\text{each batch of nodes} $\{x_i\}$ from \textbf{\texttt{SAMPLE}}($\G$)}{
\text{compute graph prompt } $z_i \leftarrow$ \text{Eq.}(\ref{eq:gnn_encoder}),\\ causal language modeling $\mathcal{L}_\text{pre-train} \leftarrow $ \text{Eq.}(\ref{eq:pre-train});\\
 \text{update} {\color{blue}{$\Theta_g$}};}
// Fine-tuning;\\
\For{\text{each batch of edges} $\{e_{ij}\}$ from \textbf{\texttt{SAMPLE}}($\G, \E_t^\text{train}$)}{
\text{compute graph prompt } $z_i, z_j \leftarrow$ \text{Eq.}(\ref{eq:gnn_encoder}),\\ $\mathcal{L}_\text{fine-tuning} \leftarrow $ \text{Eq.}(\ref{eq:fine-tune});\\
 \text{update} {\color{blue}{ $\{\Theta_g, \Theta_\text{peft}\}$}};}
// Evaluation;\\
\For{\text{each batch of nodes} $\{x_i\}$ from \textbf{\texttt{SAMPLE}}($\G$)}{
    Inference node embedding $v_i \leftarrow $ \text{Eq.}(\ref{eq:inference})
}
Compute evaluation metric on testing edges $\E_t^{test}$
\caption{Pseudo code for \Ours optimization}
 \label{alg}
\end{algorithm}

In Algorithm~\ref{alg}, we detail the training procedure of \Ours. In the pre-training phase, we iterate on every batch of nodes and sample neighbors of each node to compute the graph prompt at line 7. Line 8 and 9 correspond to the weight matrices updating of GNN prompt encoder $\Theta_g$ while freezing LLM. In the fine-tuning (the representation learning) phase, for every batch of training edges, we sample neighbors for both nodes and perform graph prompt encoding similarly. Then, we tune the parameters of PEFT and GNN prompt encoder simultaneously to minimize the representation loss (Line 14 and 15). In evaluation, we calculate the node embeddings as the \texttt{[GRL]} token in the sequence of each node and use the cosine similarity to rank the candidates edges. 

\xhdr{Parameter Efficiency.} We now describe the amount of additional parameters introduced in \Ours.
First, we use LORA~\cite{hu2021lora} as an example to calculate the PEFT parameters.
We denote the hidden dimension of GNN prompt encoder and LLM as $d_\text{GNN}$ and $d_\text{LLM}$, and rank of the LORA matrices as $r$, with $r << d_\text{model}$.
Assuming the LLM has $L$ transformer layers, LORA introduces two matrices across four components of the transformer's feedforward network: query, key, value, and output.
Consequently, the total additional parameters introduced by LoRA amount to $L \cdot 4rd_\text{model}$. 
In the case of LLaMA, where $d_\text{model}=4096$ and $L=32$, setting $r=16$ yields approximately 8.9M additional parameters in $\Theta_\text{peft}$.

For a k-layer GNN prompt encoder, setting $d_\text{GNN}=768$ introduces $k d_\text{GNN}^2\cdot N_\text{rel} + d_\text{GNN}d_\text{model}$ additional parameters, where $N_\text{rel}$ represents the number of relations in $\G$. In our experiment, with $N_\text{rel}=2,k=2$, this results in 5.5M parameters. Thus, we argue that our approach contains a comparable number of parameters to PEFT, which is significantly fewer than existing GNN-LMs (at least 110M for bert-base).

\section{Experiments} \label{sec:experiment}
To evaluate the performance of \Ours in learning representations of text-rich graphs, we focus on the following research questions (RQs) in this section:

\begin{itemize}
\item RQ1: How effective are large language models (LLMs) at graph representation learning, particularly with \Ours?
\item RQ2: Is the pre-training of the graph prompt encoder necessary?
\item RQ3: Can our approach be adapted to different PEFT methods and LLMs?
\item RQ4: Is \Ours efficient in million-scale representation learning?
\item RQ5: What is the parameter sensitivity of \Ours?
\end{itemize}
\subsection{Datasets}

\begin{table}[t]
    \centering
    \caption{Dataset Statistics.}\label{tab:dataset}
    \resizebox{\columnwidth}{!}{
    \begin{tabular}{c|ccccccc}
    \toprule
         Dataset & \#Nodes  & \#Edges & \#Task Edges & Avg Degree & Avg \#Tokens \\
    \midrule

    Clothing & 469,274 & 2,578,746 & 1792,604 & 10.99 & 117.83 \\
    Home & 453,121 & 3,732,948 & 2,382,201 & 16.48 & 133.94 \\
    Sports & 293,712 & 2,390,076 & 3,130,128 & 16.27 & 125.08 \\
    VideoGames & 38,869 & 729,062 & 844,296 & 37.51 & 147.33 \\
    \midrule
    Computer Science & 263,393 & 2,929,358 & 13,188,472 & 22.24 & 159.69 \\
    Economics & 178,670 & 3,064,144 & 3,615,524 & 34.3 & 160.21 \\
    Geology & 431,834 & 15,256,890 & 35,915,142 & 70.66 & 205.08 \\
    Mathematics & 490,551 & 4,770,644 & 12,911,644 & 19.45 & 143.94 \\
    \bottomrule
    \end{tabular}
    }
\end{table}

Some of the most popular benchmarks, such as OGB~\cite{hu2020ogb}, do not include full-text information for link prediction tasks. Following recent research trends~\cite{jin2023patton,zhang2023effect}, we benchmark on two representative text-rich graphs from the academia and e-commerce domains. In each domain, we choose four different subdomains to comprehensively study the performance of link prediction with different methods.

\xhdr{Amazon Review~\cite{mcauley2015image, he2016ups}.} The Amazon Products dataset comprises commercial data from items sold on Amazon, complete with detailed descriptions. In this dataset, items are represented as nodes in a graph, with edges connecting pairs of nodes if they are \textit{co-viewed} or \textit{co-purchased} by users. We utilize the \textit{co-viewed} edges as observed data and perform our representation learning evaluation on the \textit{co-purchased} edges, as this closely resembles real-world product recommendation scenarios. Our focus is on products within four specific subdomains: Clothing, Home, Sports, and Video Games. Each of these subdomains features a graph with over 1 million edges.

\xhdr{Microsoft Academia Graph~\cite{sinha2015overview} (MAG).} MAG is a large, heterogeneous graph containing scientific publication records, citation relationships between those publications, as well as authors, institutions, journals, conferences, and fields of study. We use a pre-processed version (MAPLE~\cite{zhang2023effect}) to obtain field specific text-rich graphs, that are Computer Science, Economics, Mathematics, Geology and \etc. We utilize the \textit{cite-by} edges as observed data and generate paper pairs that share the same author (\textit{same-author}) using author information of each paper. 

We summarize the detailed data statistics of each subdomains in Amazon Review and MAG, including average node degrees and text length in Table~\ref{tab:dataset}.

\begin{table*}[t]
\centering
\caption{Performance of link prediction on Amazon Review Graph. Each experiment is repeated three times, except for $^\dagger$: We limit API calls to just once for cost efficiency.}
\label{tab:main-exp-amazon-review}
\scalebox{0.9}{
\begin{tabular}{c|cccccccccc}
\toprule
\multirow{2}{*}{\textbf{Method}} & \multicolumn{2}{c}{\textbf{Clothing}} & \multicolumn{2}{c}{\textbf{Home\&Kitchen}} & \multicolumn{2}{c}{\textbf{Sports}} & \multicolumn{2}{c}{\textbf{Video Games}} &  \multicolumn{2}{c}{\textbf{Average}} \\ 
\cmidrule(lr){2-3} \cmidrule(lr){4-5} \cmidrule(lr){6-7} \cmidrule(lr){8-9}
& Hit@1 & MRR & Hit@1 & MRR & Hit@1 & MRR & Hit@1 & MRR & Hit@1 & MRR \\ 
\midrule

GNN (Sentence-BERT) & $74.52_{0.52}$ & $82.58_{0.4}$ & $74.18_{0.27}$ & $82.67_{0.21}$ & $61.2_{0.12}$ & $74.48_{0.16}$ & $52.64_{0.15}$ & $68.53_{0.13}$ & 65.64 &	77.07
 \\
GNN (PEFT-LLaMA) &$76.22_{0.26}$ & $84.16_{0.19}$ & $73.74_{7.03}$ & $81.66_{6.43}$ & $62.26_{0.43}$ & $75.36_{0.91}$ & $56.14_{0.18}$ & $71.59_{0.14}$ & 67.09 &	78.19\\
\midrule
Sentence-BERT &$62.11_{0.20}$ & $73.36_{0.14}$ & $65.43_{0.17}$ & $76.37_{0.14}$ & $50.13_{0.13}$ & $66.28_{0.10}$ & $41.83_{0.05}$ & $59.65_{0.10}$ & 54.88 & 68.92 \\
GraphFormers & $71.30_{0.04}$ & $79.31_{0.02}$ & $74.29_{0.03}$ & $81.63_{0.02}$ & $58.35_{0.01}$ & $71.45_{0.00}$ & $49.67_{0.04}$ & $65.98_{0.04}$ &63.40 & 74.60
\\
PATTON & $76.95_{0.02}$ & $83.79_{0.01}$ & $78.14_{0.05}$ & $84.72_{0.03}$ & $62.44_{0.02}$ & $74.62_{0.01}$ & $51.07_{0.14}$ & $66.97_{0.08}$ & 67.15 & 77.53
\\
\midrule

InstructGLM-embeddings$^\dagger$ & 76.23 & 82.60 &	79.82 & 85.93 &	62.50 &	73.25 & 48.18 & 63.00 & 66.68 &	76.20 \\
\midrule
PEFT-LLaMA & $74.73_{0.02}$ & $82.87_{0.00}$ & $78.93_{0.02}$ & $86.07_{0.02}$ & $62.52_{0.02}$ & $75.77_{0.01}$ & $56.07_{0.18}$ & $71.54_{0.14}$ & 68.06	 &79.06
 \\
GraphPEFT \textbf{w.o.} pretraining & 
$76.74_{0.07}$	&$84.57_{0.06}$	&$79.68_{0.18}$	&$86.63_{0.09}$	&$64.44_{0.14}$&	$77.21_{0.12}$	&$50.60_{0.11}$&	$66.97_{0.14}$ & 67.87 & 78.85 \\
							
GraphPEFT & \textbf{76.95}$_{{0.07}}$ & \textbf{84.71}$_{{0.06}}$ & \textbf{79.87}$_{{0.18}}$ & \textbf{86.76}$_{{0.09}}$ & \textbf{64.61}$_{{0.14}}$ & \textbf{77.34}$_{{0.12}}$ & \textbf{58.04}$_{{0.11}}$ & \textbf{73.07}$_{{0.14}}$ & \textbf{69.88} & \textbf{80.47} 
 \\
\bottomrule
\end{tabular}
}

\end{table*}
\begin{table*}[t]
\centering
\caption{Performance of link prediction on MAG graph. Each experiment is repeated three times, except for $^\dagger$: We limit API calls to just once for cost efficiency.}
\label{tab:main-exp-mag}
\scalebox{0.9}{
\begin{tabular}{c|cccccccccc}
\toprule
\multirow{2}{*}{\textbf{Method}} & \multicolumn{2}{c}{\textbf{Computer Science}} & \multicolumn{2}{c}{\textbf{Economics}} & \multicolumn{2}{c}{\textbf{Geology}} & \multicolumn{2}{c}{\textbf{Mathematics}} &  \multicolumn{2}{c}{\textbf{Average}} \\ 
\cmidrule(lr){2-3} \cmidrule(lr){4-5} \cmidrule(lr){6-7} \cmidrule(lr){8-9}
& Hit@1 & MRR & Hit@1 & MRR & Hit@1 & MRR & Hit@1 & MRR & Hit@1 & MRR \\ 
\midrule
GNN (Sentence-BERT) & $24.97_{0.15}$ & $41.18_{0.25}$ & $28.64_{0.13}$ & $43.35_{0.3}$ & $35.94_{0.22}$ & $51.28_{0.2}$ & $47.67_{0.21}$ & $62.24_{0.18}$ & 34.31 &	49.51
 \\

GNN (PEFT-LLaMA) & $27.81_{0.10}$ & $44.92_{0.13}$ & $31.89_{0.45}$ & $47.83_{0.37}$ & $35.68_{0.12}$ & $51.59_{0.13}$ & $46.24_{0.11}$ & $61.9_{0.03}$ &35.41	& 51.56
 \\
\midrule
Sentence-BERT & $20.45_{0.05}$ & $36.64_{0.03}$ & $22.65_{0.12}$ & $37.39_{0.14}$ & $29.64_{0.21}$ & $45.37_{0.13}$ & $35.85_{0.21}$ & $52.23_{0.18}$ & 27.15 &	42.91 \\

GraphFormers & $19.14_{0.09}$ & $34.08_{0.05}$ & $19.86_{0.06}$ & $32.66_{0.02}$ & $29_{0.03}$ & $43.65_{0.03}$ & $39.91_{0.13}$ & $54.50_{0.10}$ & 26.98 & 41.22\\
PATTON & $21.68_{0.12}$ & $36.48_{0.07}$ & $31.07_{0.06}$ & $44.99_{0.05}$ & $33.38_{0.13}$ & $48.06_{0.07}$ & $43.35_{0.07}$ & $57.37_{0.04}$ & 32.37 &	46.73  \\
\midrule
InstructGLM-embeddings$^\dagger$ &21.84 &	34.88& 24.76 & 36.95 & 34.55 &48.00 & 43.01 & 55.22 & 31.04 & 43.76 \\

 \midrule
 PEFT-LLaMA & $28.45_{0.62}$ & $45.16_{0.91}$ & $33.36_{0.07}$ & $49.07_{0.14}$ & $37.03_{0.47}$ & $52.56_{0.69}$ & $45.7_{0.25}$ & $61.34_{0.15}$ & 36.14	& 52.03
\\
GraphPEFT \textbf{w.o.} pretraining &$23.86_{0.19}$ & $40.11_{0.29}$ & $27.19_{0.18}$ & $41.85_{0.18}$ & $34.93_{0.25}$ & $50.37_{0.22}$ & $45.81_{0.00}$ & $60.75_{0.17}$ & 32.95 & 48.27 \\
GraphPEFT & \textbf{30.41}$_{{0.29}}$ & \textbf{47.37}$_{{0.25}}$ & \textbf{35.42}$_{{0.10}}$ & \textbf{51.08}$_{{0.13}}$ & \textbf{39.69}$_{{0.15}}$ & \textbf{55.19}$_{{0.14}}$ & \textbf{48.09}$_{{0.06}}$ & \textbf{63.02}$_{{0.01}}$ & \textbf{38.40} & \textbf{54.17}
 \\
\bottomrule
\end{tabular}
}

\end{table*}

\subsection{Baselines}
In our experiment, we compare \Ours with different state-of-the-art text-rich graph modeling algorithms (\underline{underlined} in the following paragraphs).

\xhdr{Cascaded GNN-LMs:} We consider methods that employ fine-tuned language models (LMs) or large language models (LLMs) as feature encoders, followed by training a graph neural network (GNN) on these features.  Specifically, we select MPNET~\cite{song2020mpnet} that is a variant of \underline{Sentence-BERT}~\cite{reimers2019sentence} and \underline{LLaMA} fine-tuned with LORA techniques. The LMs or LLMs are not updated in these baselines.

\xhdr{Graph-empowered LMs:} We opt to compare with \underline{GraphFormers}~\cite{yang2021graphformers}, which are designed for graph representation learning by aggregating the [CLS] tokens of neighboring nodes between transformer layers. Similar to our graph prompt approach, this method introduces virtual structure tokens to the LM.  \underline{Patton}~\cite{jin2023patton} improves upon GraphFormers by incorporating pre-training with masked token and node prediction objectives. Additionally, we also report the performance of fine-tuning Sentence-BERT~\cite{reimers2019sentence} without graph information. 

\xhdr{Graph-aware LLMs:} These methods do not alter the architectures of large language models (LLMs) but rather engage in in-context learning or fine-tuning of the LLMs. In the context of link prediction, the most suitable algorithm is \underline{InstructGLM}~\cite{ye2023natural}, which translates both the graph structure and node features into natural language. We have made minor modifications to the InstructGLM pipeline, notably by utilizing a public LLM API\footnote{text-embedding-3-small	 in \url{https://platform.openai.com/docs/guides/embeddings}} to transform the composed sequence into dense embeddings. Our adapted approach is referred to as InstructGLM-embeddings. 

At last, we design several ablations of our model to verify the effectiveness of each component: (1) \Ours without pre-training phase in Equation~\ref{eq:pre-train}; (2) \Ours without graph prompt, namely, PEFT.

For different methods, we use the same neural network architecture for GNN, LM and LLM for fair comparison. Specifically, we use
a 2-layer GraphSAGE~\cite{graphsage} as the graph neural network with hidden dimension $d_\text{GNN}$ as 768, sentence transformer (\ie, MPNet-v2~\cite{song2020mpnet}) as the pre-trained language model and LLaMA~\cite{touvron2023llama} as the default pre-trained foundation model.

\subsection{Experiment Settings}
We conduct a large-scale representation learning experiment on text-rich graphs for link prediction. Specifically, we simulate real-world scenarios supporting tasks such as item recommendation through \textit{co-purchase} prediction on Amazon Review and authorship identification through \textit{same-author} prediction on MAG.
In each subdomain, we randomly select 50,000 edges for training, 10,000 edges for validation, and use the remaining edges for testing. Following the same setup as in OGB~\cite{hu2020ogb}, we select 100 negative edges for each test edge and compute the average hit@1 and MRR score across all test edges. For all baseline methods, except GraphFormers and Patton, we employ the same contrastive loss as of Equation~\ref{eq:fine-tune}. Additionally, we standardize the neighborhood fan-outs (i.e., 5 neighbors at each hop) for GNNs. 

We implement all baselines using Deep Graph Library (\url{https://www.dgl.ai/}) and Hugging Face (\url{https://huggingface.co/}). For all methods, we train them on 50K training samples for 4 epochs on 8 Nvidia A100 GPUs with a total batch size of 32. We set the peak learning rate as 1e-4. More details can be found in Appendix B. 

\subsection{Experiment Results}
We compare our approach with state-of-the-art baselines across two different domains to assess the effectiveness of our representation learning method. We conduct three runs for each method and report the mean and standard deviation for each in Tables 1 and 2, resulting in the following observations:

\xhdr{RQ1: How effective are large language models (LLMs) at graph representation learning, particularly with \Ours?}

(1) Although masked language models (\eg, Sentence-BERT) were popularly used for sequence representation, large language models (\eg, PEFT-LLaMA) have shown improved performance over masked language models. In our experiments, LLMs yield more than 10\% improvement over Sentence-BERT.

(2) 
In the same tables, we also find GNN (Sentence-BERT) perform worse than GNN (PEFT-LLaMA). In addition, GraphFormers also reports subpar performance against GNN (PEFT-LLaMA).
Therefore, we conclude that large language models serve as powerful feature encoders for representation learning on text-rich graphs.
Meanwhile, InstructGLM outperforms GraphFormers but falls short of PATTON's performance. This discrepancy may be attributed to the challenges of handling graph structures through embedding-based in-context learning, which struggles with accurately capturing structures via natural language descriptions. Consequently, carefully fine-tuned GNN-LMs (\Ours and PATTON) tend to surpass the embeddings produced by black-box LLMs.


(3)
\Ours consistently outperforms the compared baseline across both metrics. Specifically, it surpasses PEFT-LLaMA by 2.6\% and 1.8\% on MAG and Amazon Reviews, respectively, in terms of hit@1. 
Compared to its own ablation, we find that pre-training helps improve the accuracy of link prediction, especially in the academic text-rich graph.


\xhdr{RQ2: Is the pre-training of the graph prompt encoder necessary?}

On MAG, we observe that our approach performs significantly worse ($\sim6\%$) without pretraining compared to its pretrained counterpart. On Amazon Review, while the performance gap is smaller across three subdomains, the Video Games category demonstrates instability without pretraining.
Interestingly, Patton also outperforms GraphFormers with the same architecture on subdomains that utilize pre-trained checkpoints. Both methods introduce virtual structural tokens, and employing the same pre-training objective as the pre-trained language model.
Apparently, continual training helps align the virtual token representations with text representations. Therefore, we advocate for a pre-training then fine-tuning paradigm, where one needs to pre-train only once in a domain and can fine-tune for various applications (\eg, co-purchase, churn, \etc).

\subsection{Model Analysis}
In this section, we provide more in-depth study to understand performance and efficiency of \Ours.

\xhdr{RQ3:Can our approach be adapted to different PEFT methods and LLMs?}

\begin{figure}[ht]
    \centering
    \begin{subfigure}[b]{0.22\textwidth}
        \centering
        \includegraphics[width=\textwidth]{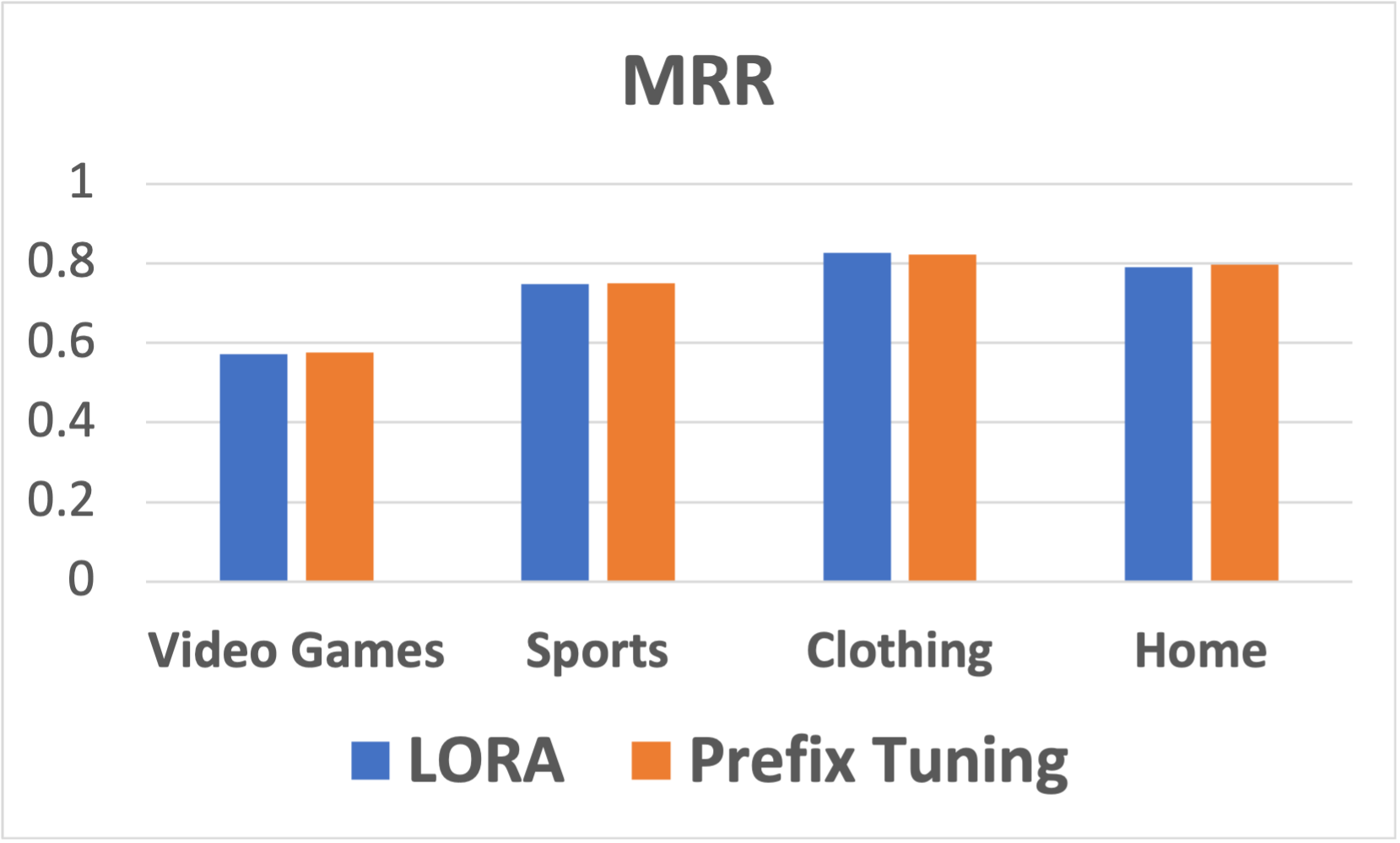}
        \caption{Varying PEFT algorithms}
        \label{fig:sub_peft}
    \end{subfigure}
    \hfill 
    \begin{subfigure}[b]{0.22\textwidth}
        \centering
        \includegraphics[width=\textwidth]{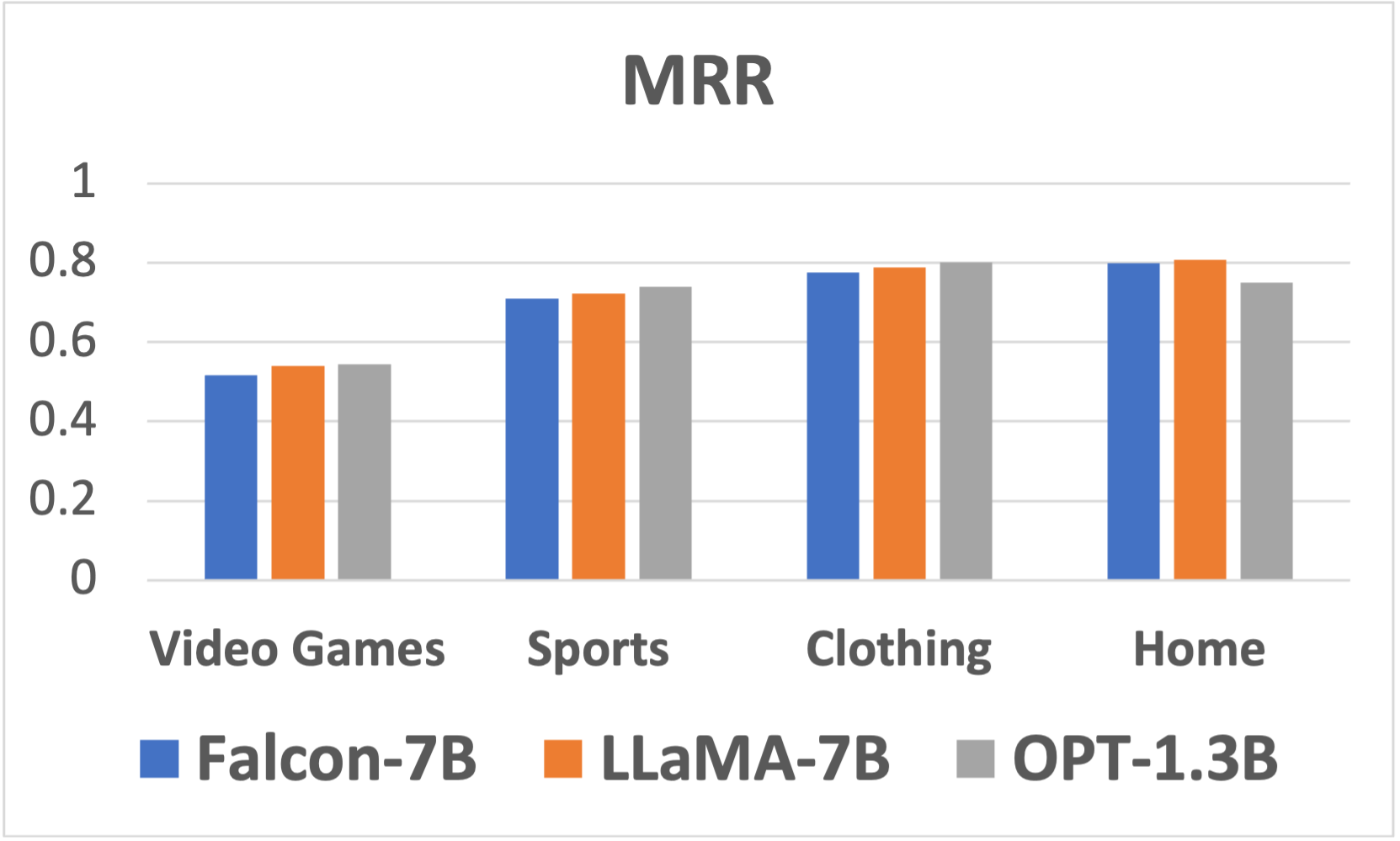}
        \caption{Varying backbone LLMs}
        \label{fig:sub_llm}
    \end{subfigure}
    \caption{\Ours using different LLMs and PEFT algorithms on Amazon Review.}
    \label{fig:verying_LLMs_study}
\end{figure}

We first apply prefix tuning~\cite{li2021prefix} on \Ours and demonstrate the results on four subdomains of Amazon Review in Figure~\ref{fig:sub_peft}. We can find that using prefix-tuning or LORA does not show much difference, which indicates the versatility of our framework on various PEFT techniques. 

Second, we substitute the backbone language model with OPT-1.3B~\cite{zhang2022opt} and Falcon-7B\footnote{\url{https://huggingface.co/tiiuae/falcon-7b}}. In Figure~\ref{fig:sub_llm}, we observe that, in general, LLaMA-7B outperforms the others, with Falcon-7B in second place and OPT-1.3B being outperformed by these two models. Our results are consistent with other work that shows LLMs with more parameters yield better performance in downstream applications.

\xhdr{RQ4: Is \Ours efficient in large-scale representation learning?}

\begin{table}[t]
\centering
\caption{
Time and memory costs of \Ours on MAG-Economics using 8 A100 GPUs with a total batch size of 32.}
\label{tab:efficiency_study}
\begin{tabular}{ c  c  c  c}
 \toprule
& pre-training & fine-tuning & inference \\
 \midrule
 Time & 36min & 96min & 11min \\
 Memory & 20883MB  & 28309MB & 13676MB   \\
 \#trainable & 5.5M & 14.4M & - \\ 
\bottomrule
\end{tabular}
\end{table}
In Table~\ref{tab:efficiency_study}, we report the running time of our approach at each phase on MAG-Economics. Specifically, unlike the pre-training time of PATTON~\cite{jin2023patton} (10hr+), the pre-training time of \Ours is even shorter than the fine-tuning time. Because we only optimize $\Theta_g$ in the pre-training phase and we can say that the pre-training is both necessary and efficient together with \textbf{RQ2}.
Owing to its parameter-efficient tuning, our approach not only minimizes the number of training parameters but also achieves the best performance. As mentioned in the introduction, the design of \Ours aims to provide a high-quality, lightweight framework suitable for various applications on different text-rich graphs. Considering the reduced parameter storage requirements and the training time, we believe \Ours holds significant potential for industrial-scale applications, including recommendation systems, ranking tasks, and \etc

\xhdr{RQ5: What is the parameter sensitivity of \Ours?}

\begin{figure}[ht]
    \centering
    \begin{subfigure}[b]{0.22\textwidth}
        \centering
        \includegraphics[width=\textwidth]{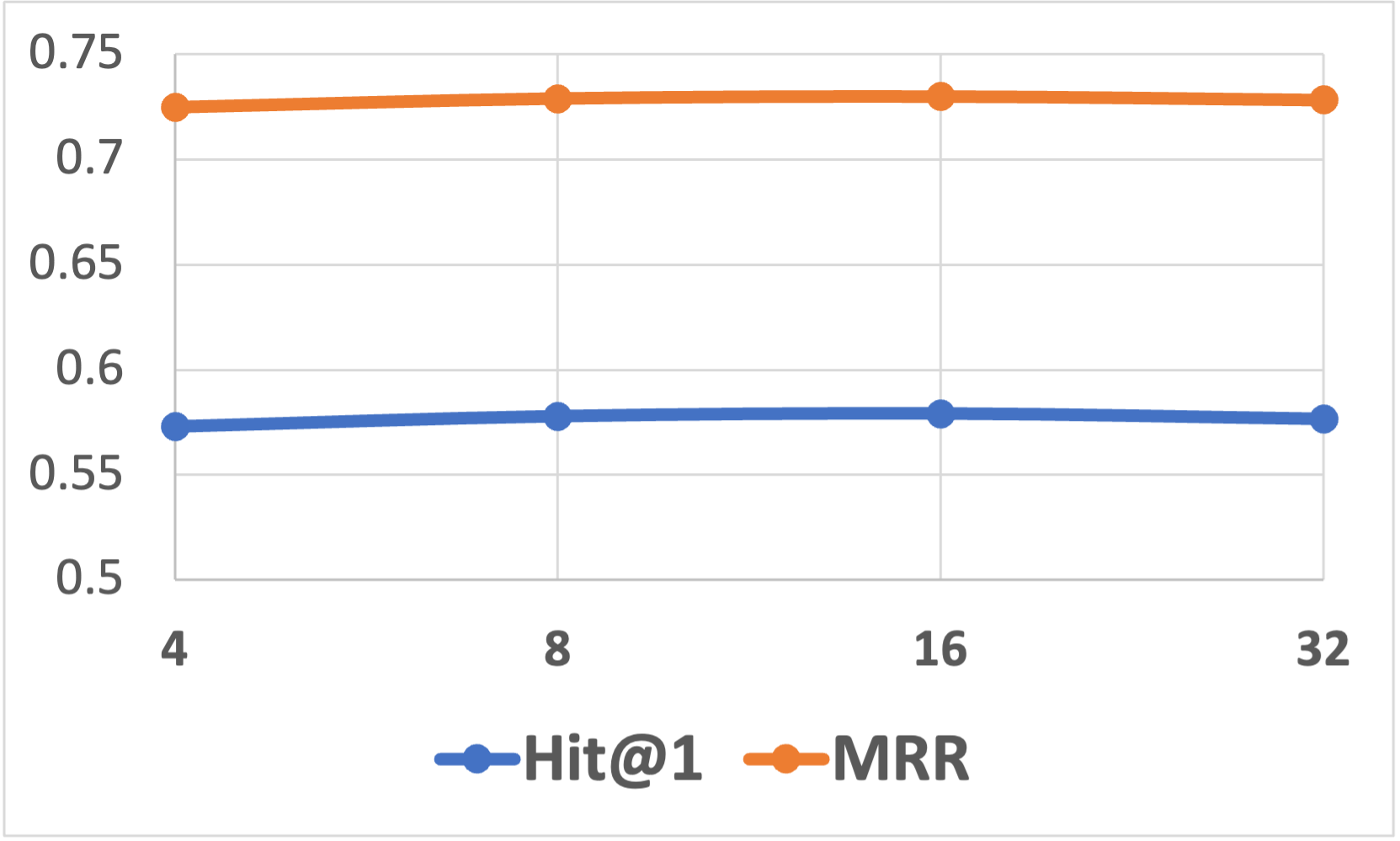}
        \caption{Varying LORA rank}
        \label{fig:sub_a}
    \end{subfigure}
    \hfill 
    \begin{subfigure}[b]{0.22\textwidth}
        \centering
        \includegraphics[width=\textwidth]{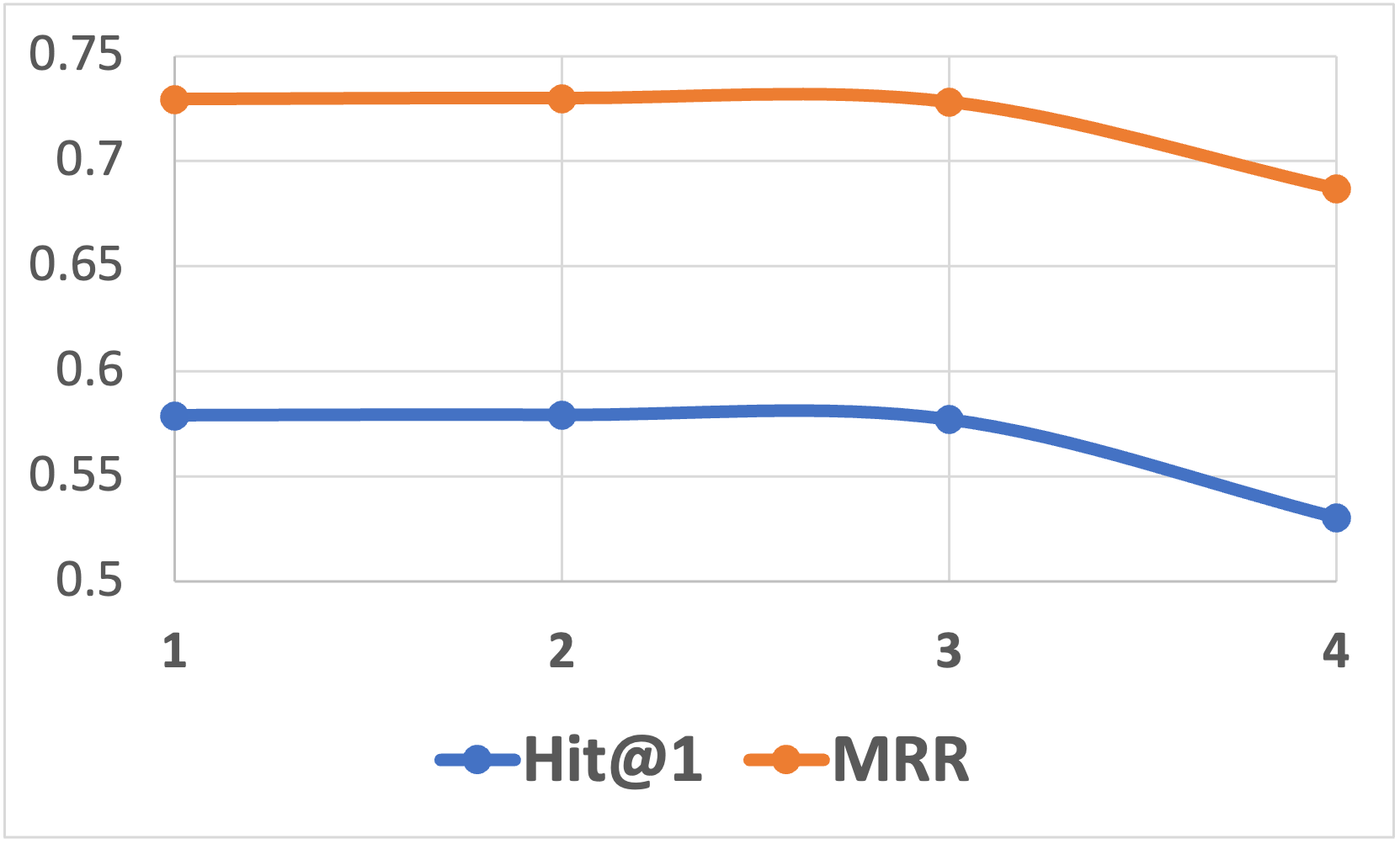}
        \caption{Varying number of hops}
        \label{fig:sub_b}
    \end{subfigure}
    \caption{\Ours using different LLMs and PEFT algorithms on Amazon Review.}
    \label{fig:param_study}
\end{figure}

There are several important hyper-parameters in \Ours: (1) LORA rank $r$, it affects the number of trainable parameters (2) number of hops $k$ in GNN prompt encoder, it affects the amount of structural information used in the training of \Ours. In Figure~\ref{fig:param_study}, we observe that varying the rank of LoRA matrices from 4 to 32 does not significantly affect the performance. This observation aligns with findings from other studies, which suggest that optimizing for a single task requires only minor adjustments to the pre-trained LLM weights. Similarly, varying number of hops in the GNN prompt encoder has a minor effect on performance, until the point where message passing begins to aggregate more noisy neighbors than useful ones (e.g., at 4 hops, as shown in Figure~\ref{fig:sub_b}).

\section{Conclusion and future work}\label{sec:conclusion}
This paper proposes \Ours to harness LLMs for representation learning on text-rich graphs. Compared with existing work of applying LLM on graph structure data, our proposed method is the first one that generates embeddings from the LLM and therefore can be applied on numerous industrial applications. More importantly, \Ours only requires a small amount of training parameters and extra storage for obtained models.
Across eight different link prediction experiments across two domains, \Ours improves hit@1 and mrr by 3\% over the second best baselines. We validate the effectiveness of \Ours with different peft methods and mutiple LLMs such as OPT, LLaMA and Falcon.

\bibliographystyle{ACM-Reference-Format}
\bibliography{reference}

\appendix
\newpage
\clearpage
\section{Theoretical Analysis} 
\begin{theorem}[]\label{thm_1}
Consider the $l$-th layer of a transformer, where $H^l_k = [h^l_{i,0}, \ldots, h^l_{i,k}]$ represents the embedding matrix for a text sequence of length $k$. At each layer, appending a graph prompt to the sequence results in a convex combination of the embeddings $[H^l_k; z_i]$. In contrast, prepending the graph prompt $z_i$ to the sequence allows for the incorporation of $z_i$ in the computation of $H^l_k$.
\end{theorem}
\begin{proof}
We first provide the calculation of two different position of graph prompt token in a casual language model. 
\paragraph{Prompt token after text sequence.}
In the $l$-th transformer layer, let us denote $H^l_j = [h^l_{i,0}, \ldots, h^l_{i,j}]$ as the embedding matrix for the first $j$ tokens. Then, the attention calculation for the $(k+1)$-th hidden states $h^{l+1}_{i,j}$, incorporating the prompt token $z_i$, is expressed as follows:
\begin{equation}
\text{Attention}(h^{l+1}_{i,j}) = \text{softmax}\left(\frac{Q(h^l_{i,j})K(H_j)^T}{\sqrt{d_k}}\right)V(H^l_j)
\end{equation}
Interestingly, in a causal language modeling setup, the attention scores for text tokens remain unchanged when a graph prompt is appended afterwards. This is because, in causal language modeling, a token does not have visibility of subsequent tokens. 
As a result, the hidden representation of $H_k^l$ is the same before and after appending the prompt token. The representation of \texttt{[GRL]} is $h_\texttt{GRL} = \textbf{a}^T [H_k^l;z_i], \sum a_j = 1$, which is a convex combination of original hidden representation and graph prompt $z_i$. This is equivalent as cascading GNN-LMs that language model and graph neural network calculations are independent with each other.

\paragraph{Prompt token before text sequence.}
When we place graph prompt in the beginning, $z_i^l$ is included in the attention score calculation of every token:
\begin{equation}
\text{Attention}(h^{l+1}_{i,j}) = \text{softmax}\left(\frac{Q(h^l_{i,j})K([z_i^l;H^l_j])^T}{\sqrt{d_k}}\right)V([z^l_i; H^l_j])
\end{equation}
In this manner, the hidden representation of text tokens, denoted as $\widehat{H_k^l} \neq H_k^l$, incorporates the information of $z_i$ because the attention mask does not exclude the prompt token. Consequently, prepending the graph prompt employs the LLM as the predictor, whereas appending the prompt utilizes the GNN as the predictor.
\end{proof}

\section{Detailed Experiment Settings}
We use the PEFT library (\url{https://github.com/huggingface/peft}) from Hugging Face to implement the LORA and Prefix Tuning of LLMs.
The graph prompt $z_i$ is inserted at the beginning of the input embedding within the forward function of Hugging Face's AutoModel, and a ``1'' is prepended to the attention mask.

The hyper-paramters of \Ours are shown in Table~\ref{tab:hyperparameter}.

\begin{table}[H]
    \caption{Hyperparameters of \Ours.}
    \label{tab:dataset-stats}
    \centering
    \scalebox{0.95}{
    \begin{tabular}{c c }
    \midrule
          parameter name  & value  \\
         \midrule 
         pre-training batch size $B$ & 8 \\
         pre-training learning rate lr & 1e-3 \\
         fine-tuning size B & 4 \\
         fine-tuning learning rate lr  & 1e-4 \\
         GNN hidden dimension $d_\text{GNN}$ & 768 \\
         LLM hidden dimension $d_\text{modem}$ & 4096 \\
         maximal sequence length N & 256 \\
         maximal gradient norm & 1.0 \\
         neighborhood size K & 5 \\
         \# GNN layers L & 2\\
         \# warmup epochs & 1 \\
         \# epochs & 4 \\
         \# random seeds & 3 \\ 
         \midrule
    \end{tabular}
    }
    \label{tab:hyperparameter}
\end{table}

\end{document}